\documentclass[conference]{IEEEtran}
\IEEEoverridecommandlockouts
\usepackage{cite}
\usepackage{amsmath,amssymb,amsfonts}
\usepackage{algorithmic}
\usepackage{graphicx}
\usepackage{textcomp}
\usepackage{xcolor}
\def\BibTeX{{\rm B\kern-.05em{\sc i\kern-.025em b}\kern-.08em
    T\kern-.1667em\lower.7ex\hbox{E}\kern-.125emX}}
\begin{document}

\title{Transformers for Multimodal Brain State Decoding: Integrating Functional Magnetic Resonance Imaging Data and Medical Metadata}

\author{\IEEEauthorblockN{1\textsuperscript{st} Danial Jafarzadeh Jazi}
\IEEEauthorblockA{\textit{Department of Quantum and Converging Sciences} \\
\textit{Islamic Azad University Central Tehran Branch}\\
Tehran, Iran \\
Danialj0081@gmail.com}
\and
\IEEEauthorblockN{2\textsuperscript{nd} Maryam Hajiesmaeili}
\IEEEauthorblockA{\textit{Department of Quantum and Converging Sciences} \\
\textit{Islamic Azad University Central Tehran Branch}\\
Tehran, Iran \\
m.hajiesmaili@yahoo.co.uk}
}

\maketitle

\begin{abstract}
    Decoding brain states from functional magnetic resonance imaging (fMRI) data is vital for advancing neuroscience and clinical applications. 
    While traditional machine learning and deep learning approaches have made strides in leveraging the high-dimensional and complex nature of fMRI data, 
    they often fail to utilize the contextual richness provided by Digital Imaging and Communications in Medicine (DICOM) metadata. This paper presents a novel framework 
    integrating transformer-based architectures with multimodal inputs, including fMRI data and DICOM metadata. By employing attention mechanisms, the proposed method captures 
    intricate spatial-temporal patterns and contextual relationships, enhancing model accuracy, interpretability, and robustness. The potential of this framework spans applications 
    in clinical diagnostics, cognitive neuroscience, and personalized medicine. Limitations, such as metadata variability and computational demands, are addressed, and future directions for 
    optimizing scalability and generalizability are discussed.
\end{abstract}


\section{Introduction}
Decoding brain states from functional magnetic resonance imaging (fMRI) data has become a cornerstone of 
modern neuroscience research, enabling insights into cognitive processes, mental health conditions, and 
neural mechanisms\cite{b1}. 
Despite its potential, analyzing fMRI data remains challenging due to its high dimensionality, temporal 
complexity, and sensitivity to noise\cite{b2}.

Traditional machine learning and deep learning approaches have made significant advancements. Yet, they 
often struggle to fully leverage the information available in fMRI datasets, particularly when metadata 
from the Digital Imaging and Communications in Medicine (DICOM) standard is included\cite{b3}.
DICOM metadata, which contains acquisition parameters, scanner characteristics, and patient demographics
, is typically underutilized in neuroimaging studies\cite{b4}.

While it is often relegated to confound correction or dataset organization, it has the potential to complement 
fMRI time-series data by providing contextual information that can improve model performance and generalizability\cite{b3, b5}.

Transformers, originally developed for natural language processing, have demonstrated exceptional performance in tasks requiring 
sequential and multimodal data processing\cite{b6}. 
Their ability to model complex relationships through attention mechanisms makes them particularly suited for analyzing the spatial-temporal 
patterns in fMRI data while simultaneously incorporating auxiliary inputs like DICOM metadata\cite{b7}.
This synergy presents an exciting opportunity to address the limitations of current brain state decoding approaches\cite{b8}.

In this paper, we propose a conceptual framework for combining transformer-based models with DICOM metadata to enhance brain state decoding from fMRI data. By leveraging the attention mechanism to integrate multimodal information, this approach aims to achieve improved robustness, interpretability, and accuracy\cite{b9}.

Although this framework is presented as a theoretical contribution, its potential impact on fields such as neuroscience, clinical diagnosis, and cognitive research is significant\cite{b1}.
The remainder of this paper is organized as follows. Section II reviews related work on brain state decoding, transformer applications in biomedical imaging, and the role of DICOM metadata. Section III introduces the proposed framework, detailing the integration of fMRI data and metadata in a transformer architecture. Section IV discusses potential advantages, limitations, and future directions. Finally, Section V concludes with a summary of key findings and implications.

\section{Related work}

\subsection{Brain State Decoding Using fMRI}
The decoding of brain states from fMRI data has been an area of extensive research,
 leveraging both traditional machine learning and deep learning methods. 
Early approaches relied on statistical methods and classical machine learning models, 
such as support vector machines (SVMs) and random forests, which demonstrated limited success in capturing the high-dimensional, 
nonlinear relationships in fMRI data\cite{b10, b11}. More recent studies have employed deep learning models, 
including convolutional neural networks (CNNs) and recurrent neural networks (RNNs), 
to extract spatial and temporal features from fMRI scans\cite{b12}. 
These approaches have achieved notable improvements in tasks such as predicting cognitive tasks, 
decoding mental imagery, and identifying psychiatric conditions\cite{b13, b14}. 
However, many of these models struggle to generalize across datasets and require extensive manual feature engineering\cite{b15, b16 }.

\subsection{Transformer-Based Models in Biomedical Imaging}
Transformers, initially developed for natural language processing, have shown remarkable potential in biomedical imaging tasks\cite{b6}.
Unlike CNNs, transformers leverage self-attention mechanisms to model long-range dependencies in data, making them particularly suited for medical image analysis\cite{b17}. 
Recent research has demonstrated the effectiveness of transformer-based architectures in segmentation, classification, and anomaly detection in modalities such as MRI, CT, and histopathological images\cite{b18, b19}. 
Vision Transformers (ViTs) and hybrid architectures combining CNNs and transformers have been shown to outperform traditional models in several benchmark datasets\cite{b20}. 
Despite these advancements, the application of transformers to fMRI data remains underexplored, particularly in the context of integrating metadata and multimodal inputs\cite{b21}.

\subsection{Use of Metadata in Medical Image Analysis}
Metadata, especially DICOM metadata, provides critical contextual information about medical images, such as acquisition parameters, patient demographics, and clinical history\cite{b22}. 
Traditional approaches have often overlooked this rich source of information, focusing exclusively on pixel-level data\cite{b23}. 
However, recent studies have highlighted the potential of incorporating metadata to improve model performance. For instance, metadata has been used to enhance the interpretability of models, correct for scanner variability, and improve diagnostic accuracy\cite{b24 , b25}. 
In multimodal frameworks, metadata could serve as an auxiliary input to bridge the gap between imaging data and clinical outcomes, yet its integration into transformer-based models remains an open challenge\cite{b21}. 

\subsection{Gaps Addressed by the Proposed Framework}
Despite significant progress in brain state decoding\cite{b26}, 
transformer-based imaging\cite{b18}, and metadata utilization\cite{b25}, 
existing studies often fail to integrate these components into a unified framework. 
Few approaches effectively combine fMRI data with metadata, limiting the ability to fully leverage multimodal information for decoding brain states\cite{b21}.
The proposed framework aims to address these gaps by designing a transformer model that integrates DICOM metadata as contextual input alongside fMRI images. 
This multimodal approach has the potential to enhance model interpretability, improve cross-dataset generalization, and provide deeper insights into brain state decoding\cite{b15, b27}.

\section{Methodology}
This research proposes a novel framework that integrates functional MRI (fMRI) data with 
DICOM metadata using a transformer-based model to improve brain state decoding\cite{b21}. 
Unlike traditional methods that focus solely on voxel-level features\cite{b23}, 
our framework leverages scanner parameters, patient demographics, and acquisition metadata encoded in DICOM files to contextualize and enrich the fMRI data “Fig. 1”\cite{b22, b25}. 
By employing a multimodal transformer architecture, the model extracts spatial-temporal patterns from fMRI data while aligning them with complementary information from metadata, 
enabling more accurate and interoperable brain state predictions\cite{b15, b18}.
\begin{figure}[t]
    \centerline{\includegraphics[width=\columnwidth]{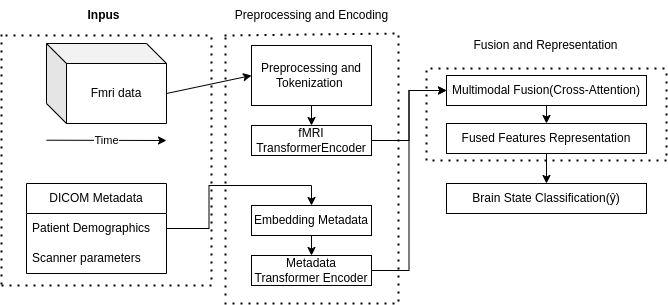}}
    \caption{Overview of the multimodal framework for brain state decoding.}
    \label{fig}
\end{figure}
\subsection{Input Modalities}
The proposed framework processes two distinct input modalities: fMRI voxel data and DICOM metadata.
\subsubsection{FMRI Data}
fMRI data captures spatial-temporal brain activity by measuring changes in blood oxygenation\cite{b27}. 
Each fMRI dataset is represented as a 4D volume (time series of 3D brain images), where voxel intensities reflect activity over time\cite{b28}. 
These data provide detailed spatial patterns (across brain regions) and temporal dynamics (e.g., neural activation sequences), forming the primary input for decoding brain states\cite{b29}.
Each dataset is represented as:
\begin{equation}
X \in \mathbb{R}^{T \times H \times W \times D},\label{eq1}    
\end{equation}
where \(T\) is the number of time points, and \(H\), \(W\), and \(D\) are the spatial dimensions (height, width, depth).
To prepare for processing:
\paragraph{\textbf{Patches}}
The 4D data is divided into patches:
\begin{equation}
    P_i \in \mathbb{R}^{t \times h \times w \times d}, \quad i = 1, 2, \dots, N,\label{eq2}
\end{equation}
where \(t\), \(h\), \(w\), and \(d\) are the dimensions of each patch, and \(N\) is the total number of patches\cite{b17}.
\paragraph{\textbf{Tokenization}}
Each patch is flattened and embedded:
\begin{equation}
    T^{\text{fMRI}}_i = f_{\text{emb}}^{\text{fMRI}}(P_i),\label{eq3}
\end{equation}
resulting in \(T^{\text{fMRI}} \in \mathbb{R}^{N \times F}\), where \(F\) is the embedding dimension\cite{b6}.

\subsubsection{DICOM Metadata}
DICOM metadata contains critical contextual information, including scanner parameters (e.g., echo time, repetition time), 
patient demographics (e.g., age, sex), and acquisition settings\cite{b22}. 
These attributes influence image quality and inter-subject variability, making them invaluable for standardizing and interpreting brain imaging data\cite{b25}. 
The metadata, represented as tabular features, complements the voxel-wise fMRI data by encoding acquisition-specific and subject-specific variability\cite{b24}.
DICOM metadata contains scanner parameters \(S\) and patient demographics \(D\)\cite{b22}. These are preprocessed as:
\begin{equation}
    M \in \mathbb{R}^{K \times F_m},\label{eq4}
\end{equation}
where \(K\) is the number of metadata attributes, and \(F_m\) is the embedding dimension for metadata\cite{b25}. For each attribute \(m_k\):
\begin{equation}
    T^{\text{meta}}_k = f_{\text{emb}}^{\text{meta}}(m_k).\label{eq}
\end{equation}
The use of metadata embeddings enhances the interpretability and performance of the model\cite{b24}.

\subsection{Framework}
The proposed framework utilizes a multimodal transformer model to jointly process fMRI voxel data and DICOM metadata. 
This approach leverages the strengths of transformer architectures to capture complex relationships within and across modalities, 
enabling more accurate and interpretable brain state decoding.
\subsubsection{Embedding Layers}
The fMRI data \(X_{\text{fMRI}}\) is first tokenized into non-overlapping 3D patches, 
each represented as a vector \(\mathbf{p}_i \in \mathbb{R}^d\), where \(d\) is the embedding dimension. 
These patches are embedded into a continuous vector space using a learnable embedding matrix \(\mathbf{W}_e\) and positional encoding \(\mathbf{e}_{\text{pos}}(i)\) to retain spatial and temporal information:
\begin{equation}
    \mathbf{z}_i^{(0)} = \mathbf{W}_e \cdot \mathbf{p}_i + \mathbf{e}_{\text{pos}}(i),\label{eq6}
\end{equation}
where \(\mathbf{W}_e \in \mathbb{R}^{d \times p}\) (with \(p\) being the flattened patch size) and \(\mathbf{e}_{\text{pos}}(i)\) is a sinusoidal positional encoding\cite{b6}.
For metadata \(X_{\text{meta}}\), each attribute vector \(\mathbf{x}_i\) (e.g., scanner parameters, patient demographics) is embedded into the same vector space as the fMRI tokens:
\begin{equation}
    \mathbf{m}_i^{(0)} = \mathbf{W}_m \cdot \mathbf{x}_i,\label{eq7}
\end{equation}
where \(\mathbf{W}_m \in \mathbb{R}^{d \times f}\) (with \(f\) being the number of metadata features) is the metadata embedding matrix. 
Categorical attributes (e.g., sex, scanner type) are one-hot encoded before embedding, while numerical attributes (e.g., age, echo time) 
are normalized\cite{b25}.
\subsubsection{Transformer Architecture}
The model applies a stack of transformer encoder layers to process both modalities. 
Each encoder layer consists of multi-head self-attention and feedforward layers. 
The self-attention mechanism computes query (\(\mathbf{Q}\)), key (\(\mathbf{K}\)), and value (\(\mathbf{V}\)) matrices as:
\begin{equation}
    \mathbf{Q} = \mathbf{W}_q \cdot \mathbf{Z}, \, \mathbf{K} = \mathbf{W}_k \cdot \mathbf{Z}, \, \mathbf{V} = \mathbf{W}_v \cdot \mathbf{Z},\label{eq8}
\end{equation}
where \(\mathbf{Z} = [\mathbf{Z}_{\text{fMRI}}; \mathbf{Z}_{\text{meta}}]\) represents the concatenated fMRI and metadata embeddings, 
and \(\mathbf{W}_q\), \(\mathbf{W}_k\), \(\mathbf{W}_v\) are learnable weight matrices. 
The attention output is computed as:
\begin{equation}
    \text{Attention}(\mathbf{Q}, \mathbf{K}, \mathbf{V}) = \text{softmax}\left(\frac{\mathbf{Q} \cdot \mathbf{K}^\top}{\sqrt{d_k}}\right) \cdot \mathbf{V},\label{eq9}
\end{equation}
where \(d_k\) is the dimension of the key vectors\cite{b6}.

To fuse information from both modalities, the model incorporates cross-attention layers. 
The cross-attention mechanism computes:
\begin{equation}
    \mathbf{Z}_{\text{fMRI}}^{(l+1)} = \text{Attention}(\mathbf{Q}_{\text{fMRI}}, \mathbf{K}_{\text{meta}}, \mathbf{V}_{\text{meta}}),\label{eq 10}
\end{equation}
where \(\mathbf{Q}_{\text{fMRI}}\), \(\mathbf{K}_{\text{meta}}\), 
and \(\mathbf{V}_{\text{meta}}\) are derived from the fMRI and metadata embeddings, 
respectively. This allows the model to attend to relevant metadata features while processing fMRI 
data\cite{b17}.
\subsubsection{Output Layer}
The fused representation \(\mathbf{Z}_{\text{fused}}\) is passed through a classification head to predict brain states. A softmax layer computes the probabilities for each state:
\begin{equation}
    \hat{y} = \text{softmax}(\mathbf{W}_{\text{out}} \cdot \mathbf{Z}_{\text{fused}}),\label{eq 11}    
\end{equation}
where \(\mathbf{W}_{\text{out}} \in \mathbb{R}^{C \times d}\) (with \(C\) being the number of brain states) is the output weight matrix.
\subsection{Training and Optimization}
The framework is trained using a composite loss function that combines \textbf{cross-entropy loss} 
for brain state classification and \textbf{domain-adaptive loss} 
to minimize dataset variability.
\subsubsection{Cross-Entropy Loss}
For brain state classification, the cross-entropy loss is defined as:
\begin{equation}
    \mathcal{L}_{CE} = - \sum_{i=1}^C y_i \log(\hat{y}_i),\label{eq 12}    
\end{equation}
where:
\begin{itemize}
    \item \(C\) is the number of classes,
    \item \(y_i\) is the ground truth label for class \(i\),
    \item \(\hat{y}_i\) is the predicted probability for class \(i\), computed by the softmax layer in the output layer.
\end{itemize}

\subsubsection{Total Loss}
The total loss is a weighted combination of the cross-entropy loss and the domain-adaptive loss:
\begin{equation}
    \mathcal{L} = \mathcal{L}_{CE} + \lambda \mathcal{L}_{DA},\label{eq 13}
\end{equation}
where \(\lambda\) is a hyperparameter that controls the contribution of the domain-adaptive loss. 
The value of \(\lambda\) is chosen through cross-validation or empirical tuning.

\subsubsection{Optimization}
The model is optimized using the \textbf{AdamW optimizer} with a learning rate scheduler\cite{b31}. 
Regularization techniques such as \textbf{dropout}\cite{b32} and \textbf{weight decay} are employed to reduce overfitting.

\subsection{Challenges}
Developing a framework that integrates fMRI data with DICOM metadata presents several challenges that must be addressed to ensure its efficacy and scalability. 
These include:
\subsubsection{Multimodal Data Alignment}
One of the primary challenges is aligning the spatial-temporal features of fMRI data with the contextual information from DICOM metadata. 
Temporal mismatches or missing metadata can introduce noise and misalignments, potentially degrading model performance. 
Strategies such as temporal interpolation, imputation techniques for missing metadata, and dynamic alignment algorithms are required to bridge these differences effectively\cite{b33}.

\subsubsection{Heterogeneity in Metadata}
DICOM metadata often varies across scanners, institutions, and acquisition protocols. This variability can hinder the consistency of metadata embeddings and model generalizability. 
Standardization techniques, such as harmonization of metadata attributes and normalization across datasets, 
are crucial to mitigate this issue. Additionally, incorporating domain adaptation techniques could help handle diverse data sources\cite{b30}.

\subsubsection{High Dimensionality and Computational Costs}
fMRI data is inherently high-dimensional, with each scan comprising thousands of voxels over time. When combined with metadata embeddings, 
the computational demands of the transformer model increase significantly. This challenge necessitates efficient data processing pipelines, 
including dimensionality reduction methods, token pruning, or lightweight transformer variants to balance computational efficiency and model performance\cite{b6}.

\subsubsection{Overfitting in Multimodal Models}
The increased complexity of combining two data modalities (fMRI and metadata) heightens the risk of overfitting, 
especially when training on smaller datasets. Regularization techniques, such as dropout, data augmentation, 
or pretraining on large-scale datasets, could be employed to combat this issue and improve generalization\cite{b32}.

\subsubsection{Interpretability of Multimodal Models}
While transformers are powerful in capturing complex patterns, their black-box nature can pose challenges in interpreting the contributions of fMRI features and metadata attributes to the predictions. 
Incorporating explainability techniques, such as attention visualization or Shapley value analysis, can improve transparency and foster trust in the framework's outputs\cite{b34}.

Addressing these challenges will be critical in realizing the potential of this framework and ensuring its adoption in real-world applications.

\section{Discussion and future works}
\subsection{Strengths}
The proposed framework offers several strengths, particularly in integrating multimodal data for brain state decoding. 
By combining the spatial-temporal features of fMRI data with the contextual richness of DICOM metadata, it ensures a more holistic understanding of brain activity\cite{b1}. 
The transformer-based architecture excels in handling complex relationships between these data modalities through self-attention and cross-attention mechanisms, 
which allow the model to dynamically prioritize relevant features from each modality\cite{b6}. This multimodal integration improves robustness, 
as the metadata provides crucial contextual information (e.g., scanner parameters or patient demographics) that complements the variability inherent in fMRI signals\cite{b3}. 
Furthermore, by incorporating metadata, the framework enhances interpretability. Metadata attributes, such as age or acquisition parameters, can provide insights into model predictions, fostering transparency and clinical trust\cite{b4}. 
The flexibility of the transformer model also makes it adaptable to diverse datasets, supporting broader applicability in neuroimaging studies\cite{b17}.

\subsection{Limitations}
Despite its strengths, this framework faces theoretical and practical challenges. A primary limitation is the reliance on the quality and completeness of DICOM metadata. 
Missing or inconsistent metadata can affect model performance and introduce biases\cite{b5}. 
Additionally, the high dimensionality of fMRI data and the computational intensity of transformer models pose scalability concerns, 
particularly when applied to large datasets or in resource-limited settings\cite{b7}. 
Another challenge lies in the variability of acquisition protocols and scanner configurations, which may reduce the generalizability of the framework across different datasets\cite{b8}. 
Furthermore, while attention mechanisms enhance interpretability, they may not fully capture the causal relationships between fMRI signals and metadata attributes, limiting the depth of insights into brain mechanisms\cite{b9}.

\subsection{Future Work}
To address these limitations and expand the utility of the framework, several future directions can be pursued. First, implementing metadata imputation techniques and harmonization pipelines can help manage incomplete or inconsistent metadata, improving data quality and model reliability\cite{b4}. 
Additionally, optimizing the transformer architecture through lightweight or hybrid models can reduce computational costs, making the framework more scalable and accessible for real-world applications\cite{b18}. 
For validation, rigorous benchmarking against state-of-the-art methods on diverse datasets is essential to assess the framework’s robustness and generalizability [10]. Cross-institutional studies, incorporating data from multiple scanner types and patient populations, would further validate its applicability\cite{b13}.

\section{Conclusion}
Finally, this framework holds significant potential for clinical applications, such as personalized treatment planning for neurological disorders or early detection of conditions like Alzheimer’s disease\cite{b14}. 
By leveraging metadata-driven insights, clinicians could better contextualize fMRI findings, enabling more precise and individualized interventions\cite{b15}. Expanding the framework to include other imaging modalities, such as EEG or PET scans, represents another exciting avenue for future research, 
further enhancing its scope and clinical relevance\cite{b16}.

\end{document}